\documentclass[11pt]{article}

\usepackage[preprint]{acl}

\usepackage{times}
\usepackage{latexsym}
\usepackage{booktabs}

\usepackage{amsmath}
\usepackage{amssymb}
\usepackage[most]{tcolorbox}
\usepackage{multirow}
\usepackage{pifont}

\usepackage[T1]{fontenc}

\usepackage[utf8]{inputenc}

\usepackage{microtype}

\usepackage{inconsolata}

\usepackage{graphicx}

\title{An Empirical Study of Counterfactual Self-Explanations in LLMs}

\author{
  \textbf{Giannis Kalyvas\textsuperscript{1}},
  \textbf{Giorgos Filandrianos\textsuperscript{1}},
  \textbf{Orfeas Menis Mastromichalakis\textsuperscript{2}},
  \\
  \textbf{Vassilis Lyberatos\textsuperscript{1}},
  \textbf{Giorgos Stamou\textsuperscript{1}}
\\
\\
  \textsuperscript{1} National Technical University of Athens, Athens, Greece
  \\
  \textsuperscript{2} Instituto de Telecomunicações, Lisbon, Portugal
\\
  \small{
    \href{mailto:gian.kalyva@gmail.com}{gian.kalyva@gmail.com}
  }
}
\begin{document}
\maketitle
\begin{abstract}

Large language models can easily generate explanations for their own outputs, but such self-explanations are not necessarily faithful to the model’s behavior. We study this issue through counterfactual self-explanations, where a model minimally edits an input so that its own prediction changes. Across sentiment analysis and natural language inference, we evaluate ten instruction-tuned models from the LLaMA-3 and Qwen-2.5 families, measuring faithfulness, minimality, and alignment with human-annotated rationales. Our results show that model scale is the strongest determinant of explanation quality: larger models are substantially more likely to generate counterfactuals that flip their own predictions and target decision-relevant evidence. In contrast, the rationale-guided condition produces edit-minimal counterfactuals that are also more human-aligned. However, it does not consistently improve faithfulness. Overall, counterfactual self-explanations can provide useful behavioral evidence about model decisions, but their reliability depends strongly on model capacity and should be empirically validated rather than assumed\footnote{Code available at: https://github.com/gianniskalyvas/cf-self-explanations}.

\end{abstract}

\section{Introduction}\label{sec:introduction}

Large language models (LLMs) are increasingly asked not only to produce answers, but also to justify them. These self-explanations are attractive because they are easy to obtain, expressed in natural language, and often appear convincing to users. However, a fluent explanation is not necessarily a faithful one: a model may provide a plausible post-hoc rationale that does not reflect the factors that actually determine its behavior \cite{atanasova-etal-2023-faithfulness, jacovi-goldberg-2020-towards, siegel-etal-2024-probabilities}. This raises an important question for interpretability: when, if ever, can self-generated explanations be treated as reliable evidence about a model's own predictions?

We study this question through \emph{counterfactual self-explanations}. In this setting, the same model first makes a prediction and is then asked to minimally edit the input so that its own prediction changes. Counterfactuals are useful because they connect explanation quality to observable model behavior: a faithful counterfactual should cross the model's decision boundary, while human alignment reveals whether the model’s decision logic mirrors human reasoning. Together, faithfulness and human alignment distinguish merely superficial explanations from cases where the model both captures its own behavior and relies on evidence that correlates with human reasoning.

Prior work has examined LLM-generated counterfactual explanations and raised concerns about their reliability. Existing studies evaluate whether such explanations are behaviorally faithful across settings \citep{madsen2024self, atanasova-etal-2023-faithfulness, nguyen2024llmsgeneratingevaluatingcounterfactuals, randl2025mind, mayne-etal-2025-llms}, how prompting affects their generation \citep{bhattacharjee2405zero, dehghanighobadi2025can, 10.5555/3666122.3669397, chen2025reasoning}, and how factors such as intervention consistency, scale, verbosity, and counterfactual behavior shape self-explanations \citep{chuang2026faithlm, siegel2025verbosity, hanrfeval, hong2026llm}. However, these works do not jointly examine whether counterfactual self-explanations both reflect the model's own behavior and align with human-relevant evidence. 

In this work, we ask what determines whether counterfactual self-explanations are faithful, minimal, and aligned with decision-relevant evidence. Specifically, we examine how model scale affects the ability of LLMs to generate counterfactuals that flip their own predictions, whether faithful counterfactuals modify evidence that aligns with human-annotated rationales, and whether prompting choices improve self-explanations. We conduct an empirical study across sentiment analysis and natural language inference using ten instruction-tuned open-weight models from the LLaMA-3 and Qwen-2.5 families. Evaluating counterfactual explanations is inherently multifaceted, as their desirable properties can depend on both the generation method and the intended explanatory objective \citep{filandrianos-etal-2023-counterfactuals, mastromichalakis2025beyond}. We therefore evaluate counterfactual quality along three dimensions: faithfulness, measured by whether the generated edit flips the model's own prediction; minimality, measured by closeness to the original input; and human alignment, measured by overlap with human-annotated rationales. To capture the latter, we introduce Evidence-Supported Modification Precision (ESMP), which measures whether the model edits human-supported evidence rather than arbitrary input tokens.

\begin{figure*}[!h]
  \includegraphics[width=\linewidth]{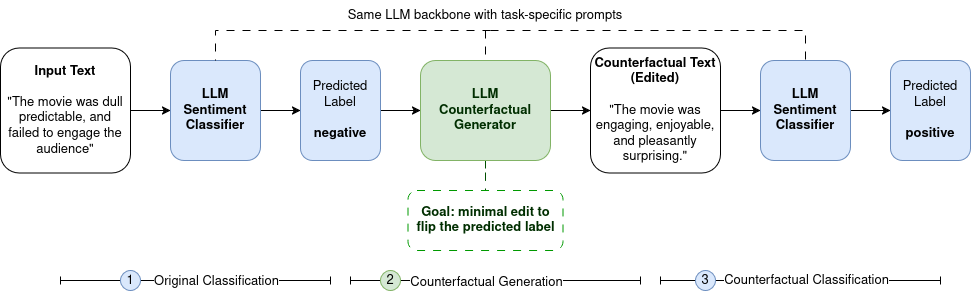}
  \caption{Experimental pipeline.}
  \label{fig:pipeline}
\end{figure*}

\section{Methodology}

We evaluate our framework on two tasks: sentiment analysis on movie reviews and natural language inference on e-SNLI. These tasks allow us to examine counterfactual self-explanations across different forms of language understanding. Both tasks are evaluated using the corresponding ERASER benchmark \cite{deyoung-etal-2020-eraser} test sets. For Movie Reviews, ERASER uses the dataset from \cite{zaidan-eisner-2008-modeling}, while for SNLI, ERASER uses e-SNLI dataset from \cite{NEURIPS2018_4c7a167b}. To assess alignment with human reasoning, we use human rationale annotations provided by ERASER as external evidence for whether model edits focus on text regions humans consider relevant for the prediction.

To isolate model-scale effects, we compare models only within the same architectural family. We evaluate ten models from two open-weight families, LLaMA-3 \citep{grattafiori2024llama3herdmodels} and Qwen-2.5 \citep{qwen25technicalreport}, ranging from 1B to 70B+. These families offer broad size ranges, consistent architectures, and public instruction-tuned variants, which help ensure prompt adherence. Further details on models, inference hyperparameters, and hardware are provided in Appendix~\ref{app:model-ids}.

\subsection{Counterfactual Generation Pipeline}

The counterfactual generation pipeline, shown in Figure~\ref{fig:pipeline} and adapted from \cite{madsen2024self}, follows a three-stage process using the same LLM with role-specific prompts. In the first stage, the model is prompted to perform the original task and produce a label for the given input, which serves as the reference point. In the second stage, the model generates a counterfactual by minimally modifying the original text such that the predicted label would flip. In the third stage, the generated counterfactual is fed back into the model for re-classification.

\subsection{Different Prompting Methods}

We vary the prompting strategy used in counterfactual generation to examine how different forms of guidance influence the resulting counterfactuals. Specifically, we consider three variants. In the baseline setting, counterfactuals are generated by conditioning the model on the opposite of the label predicted in Stage 1. We directly instruct models to generate a minimally edited counterfactual that would be classified as the target label without additional structure or guidance. In the Chat-History setting, Stages 1 and 2 are conducted within a single dialogue context. In the Rationale-Guided setting, Stage 2 is decomposed into two steps following \citep{bhattacharjee2405zero}: the model first identifies the key rationale underlying the original prediction and then generates a counterfactual by modifying these critical elements. See Appendix~\ref{app:prompts} for the prompts used.

\subsection{Evaluation Protocol}

Counterfactual explanations describe how an outcome would change under a minimal alteration to the input \citep{molnar2020interpretable}. 
Formally, a counterfactual $x'$ for input $x \in \mathcal{X}$, given a model $f : \mathcal{X} \to \mathcal{Y}$ and distance $d : \mathcal{X} \times \mathcal{X} \to \mathbb{R}_{\geq 0}$, is defined as the solution to the optimization problem:

\vspace{-1.5em}
\begin{equation}
\label{equation:counterfactual-definition}
x' = \arg\min_{x' \in \mathcal{X}} d(x, x')
\quad \text{s.t.} \quad f(x') \neq f(x).
\end{equation}
\vspace{-1em}

We evaluate two objectives derived from this definition: faithfulness and minimality. Following prior work on textual counterfactual generation, we use minimality in the empirical edit-minimality sense: a counterfactual is more minimal when it preserves more of the original input. We operationalize this objective through closeness-based metrics, including normalized edit distance and semantic similarity. This should not be interpreted as a claim that generated counterfactuals are globally minimal over the full natural-language input space. Additionally, we introduce a third metric that measures the extent to which the modifications align with human-annotated evidence.

\paragraph{Faithfulness}

The primary objective of a counterfactual is to induce a change in the model prediction. Let $f(\cdot)$ denote the classifier, $x$ the original input, and $x'$ the generated counterfactual instance.
We measure faithfulness as in \cite{madsen2024self} as the flip rate:

\vspace{-1.5em}
\begin{equation}
\text{Faithfulness} =
\frac{1}{N} \sum_{i=1}^{N}
\mathbf{1}\big[ f(x_i') \neq f(x_i) \big]
\end{equation}
\vspace{-1em}

where $N$ is the number of instances and $\mathbf{1}[\cdot]$ indicates successful label flipping.

\paragraph{Closeness}

The second objective, derived from Equation~\ref{equation:counterfactual-definition}, is to minimize the distance $d(x, x')$ between the original input and the counterfactual. Since textual similarity cannot be captured by a single distance measure, we consider multiple instantiations of $d$.

We first consider closeness as a measure of surface-level similarity, defined as the complement of the normalized edit distance:

\vspace{-1.5em}
\begin{equation}
\text{Closeness}(x, x') = 1 - \frac{\text{editdistance}(x, x')}{\max(|x|, |x'|)}
\end{equation}
\vspace{-1em}

We then consider semantic similarity, measured as the cosine similarity between sentence embeddings produced by an MPNet encoder, capturing meaning preservation beyond lexical overlap.

\paragraph{Evidence-Supported Modification Precision}

While faithfulness and minimality are standard evaluation axes in prior work, they do not capture whether the model modifies decision-relevant parts of the input. To address this, we introduce ESMP, a metric that leverages the human-provided annotations in ERASER to assess whether model edits align with annotated rationales. 

We first align the original input and the counterfactual via minimal-edit sequence alignment (replacements, deletions, insertions), identifying the set of edits needed to transform the original into the counterfactual. Replacements, deletions, or insertions that fall inside annotated evidence spans are considered true positive (TP), while the remaining, unsupported edits are considered false positive (FP).
ESMP is defined as:

\vspace{-0.5em}
\begin{equation}
\text{ESMP} = \frac{TP}{TP + FP}
\end{equation}
\vspace{-1em}

We use precision rather than recall because counterfactual explanations are intended to be minimal, so ESMP measures whether the modifications made are concentrated on human-annotated evidence, rather than measuring what fraction of all evidence is covered (a recall-style objective would instead reward broader edits, conflicting with minimality).

\section{Results and Discussion}
\label{sec:results}

We evaluate self-explanations by faithfulness, minimality, and human alignment: whether edits flip the model's prediction, remain close to the input, and overlap with human rationales. Detailed results are reported in Appendix~\ref{app:detailed-results}.

\begin{figure}[!h]
    \centering
    \includegraphics[width=1\linewidth]{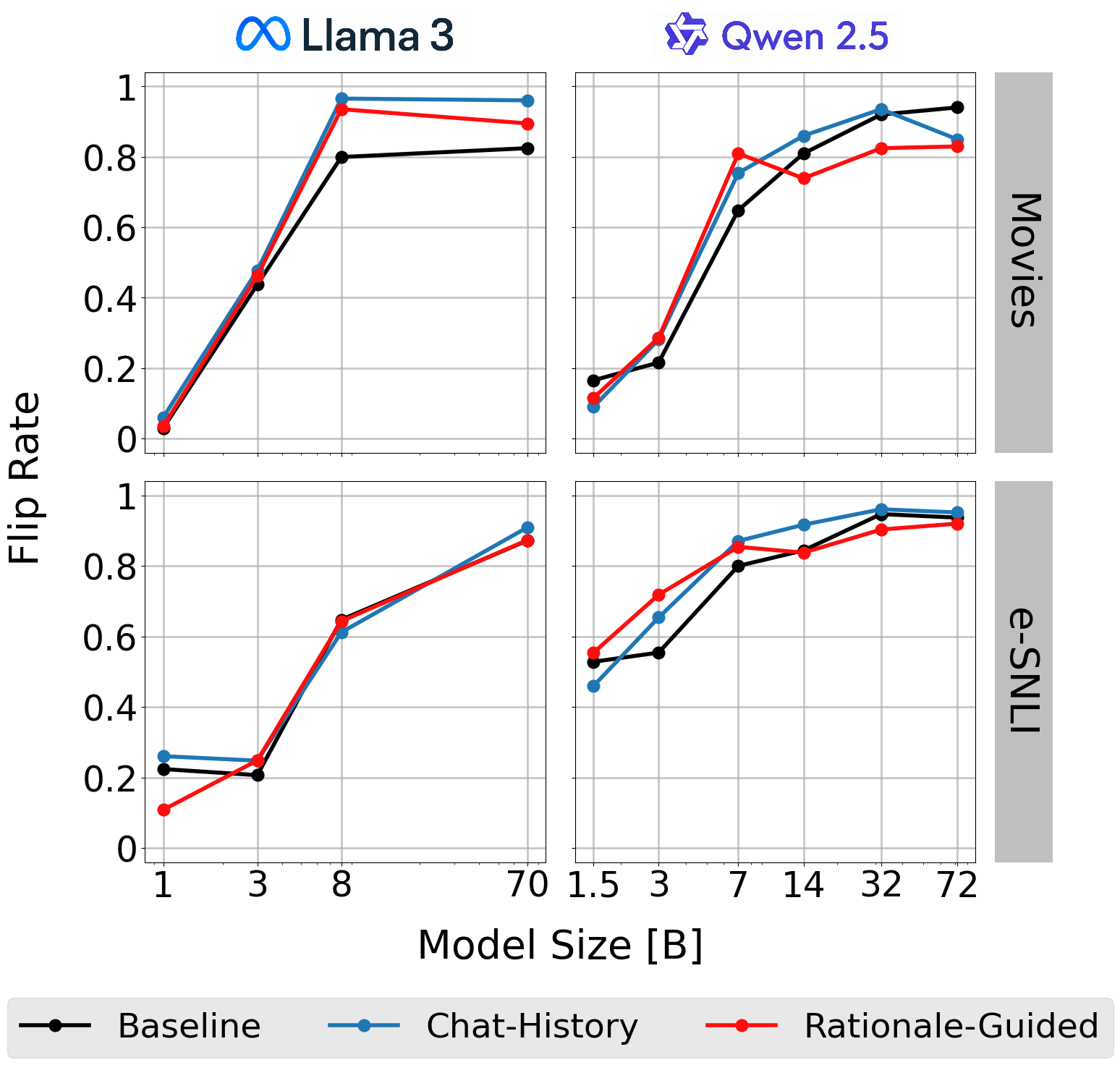}
    \caption{Faithfulness evaluation, measured by flip rate. Model size is a key factor in producing faithful self-explanations.}
    \label{fig:faithfulness}
\end{figure}

\textbf{Larger models provide more faithful self-explanations.}
Across tasks, model families, and prompting strategies, larger models are consistently better at producing self-explanations that change their own predictions (Figure~\ref{fig:faithfulness}). This suggests that self-explanation is not only a generation problem, but also a capacity problem: the model must identify which input parts support its decision and modify them in a way that crosses its own decision boundary. To quantify this trend, we compute the correlation between model size and each metric across all experimental settings. Model size is strongly correlated with flip rate ($\rho=0.87$), showing that larger models are much more likely to produce behaviorally faithful self-explanations. Methodological details and aggregate results are reported in Appendix~\ref{app:aggregate-analysis}. A regression analysis controlling for dataset, model family, and prompting strategy leads to the same conclusion: each doubling in model size is associated with an increase of 12 percentage points in self-explanation faithfulness. Thus, scale is not a minor variation across rows in the table, but a systematic driver of whether models can explain their own predictions through effective edits.

\begin{figure}[!h]
    \centering
    \includegraphics[width=1\linewidth]{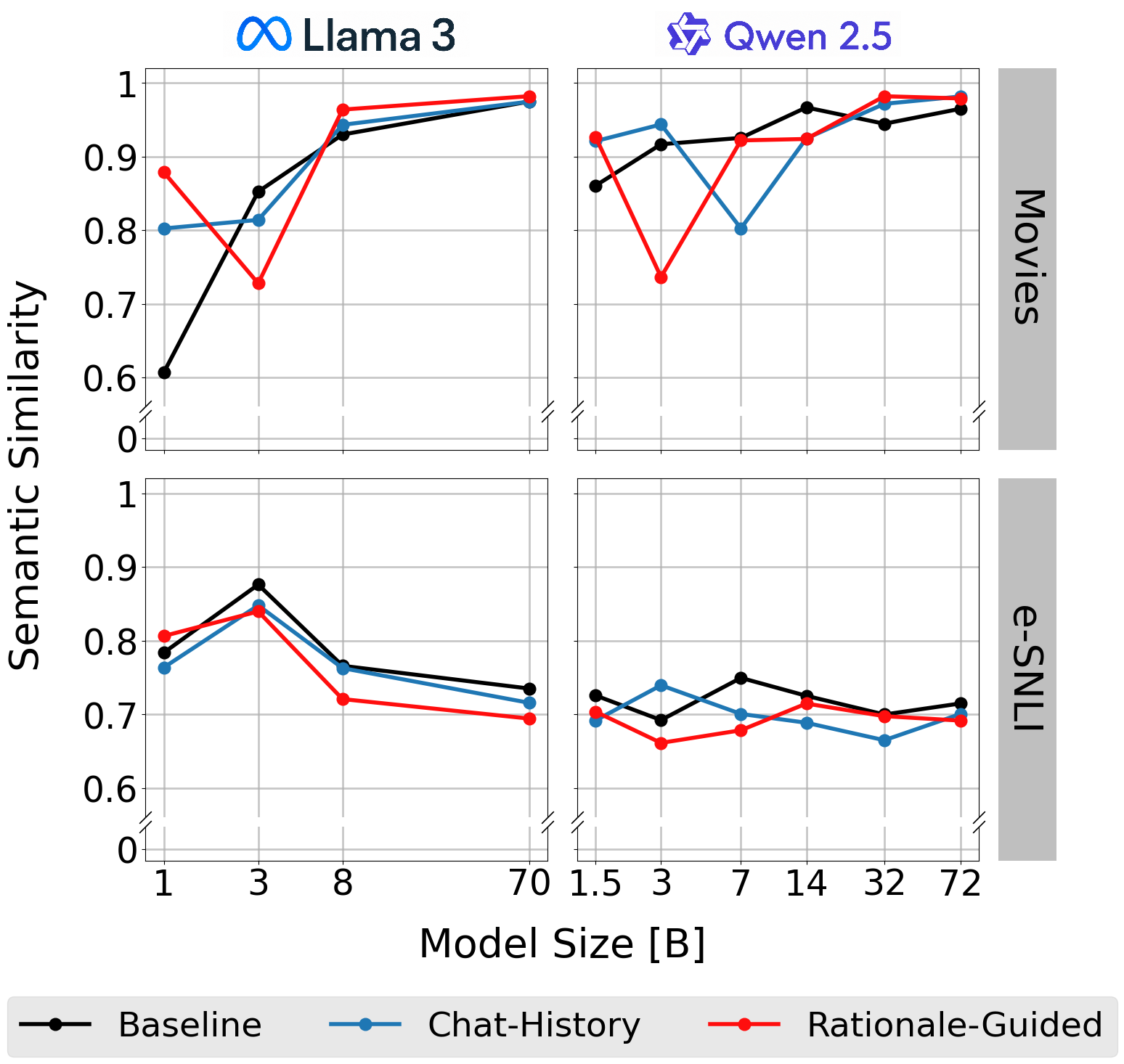}
\caption{Semantic similarity evaluation. Self-explanations remain semantically close to the original input across all settings.}
\label{fig:minimality}
\end{figure}

\begin{figure}[!h]
    \centering
    \includegraphics[width=1\linewidth]{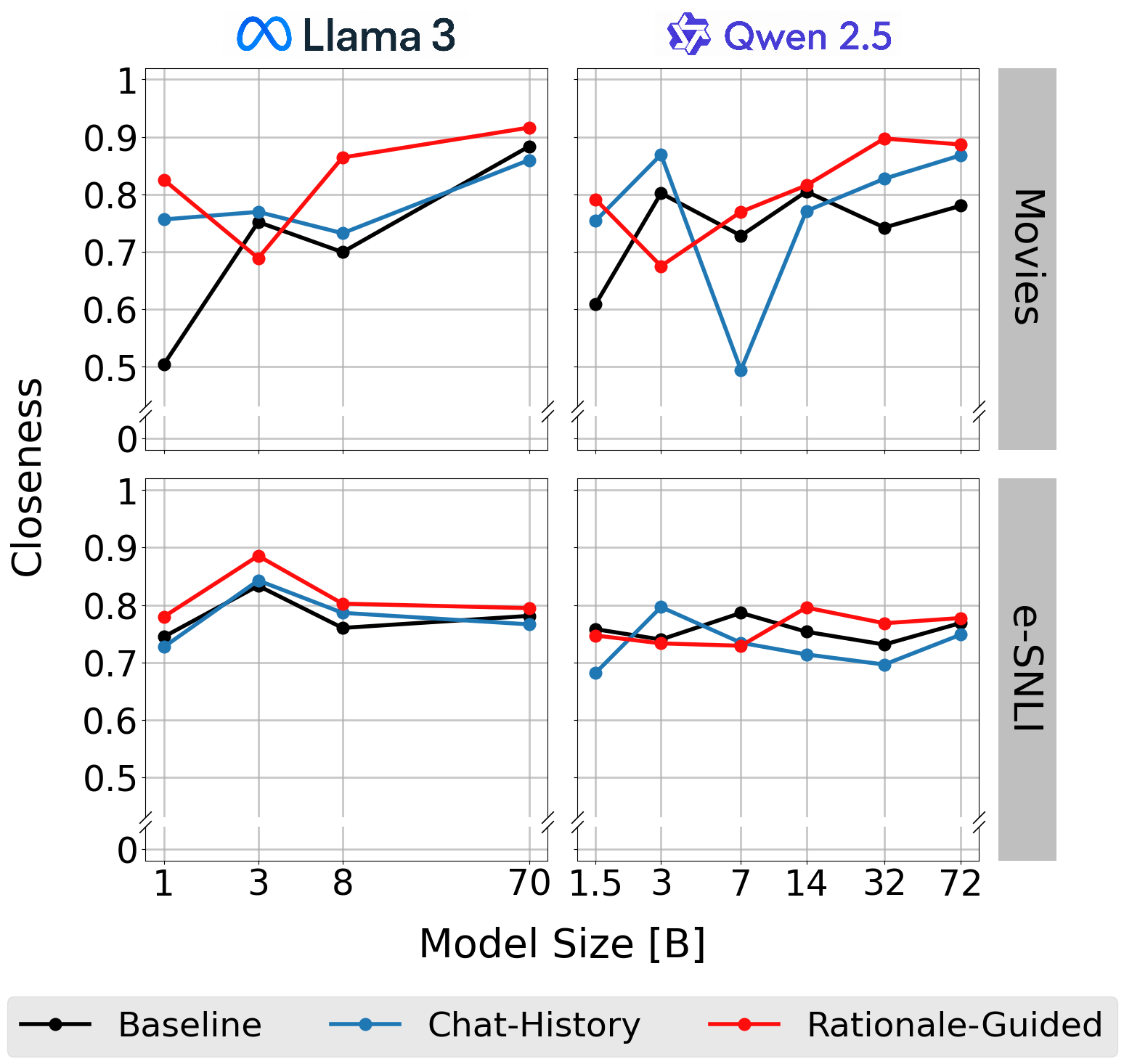}
    \caption{Closeness evaluation. The generated counterfactuals are consistently close to the input across all settings.}
    \label{fig:closeness}
\end{figure}

\textbf{The generated counterfactuals remain close to the original inputs.} Figures~\ref{fig:minimality} and \ref{fig:closeness} report the closeness scores across datasets, model families, model sizes, and prompting strategies. Overall, the generated counterfactuals remain close to the original inputs across most settings, indicating that the models generally perform localized edits rather than broad rewrites. This supports the minimality of the generated self-explanations.

\begin{figure}[!h]
    \centering
    \includegraphics[width=1\linewidth]{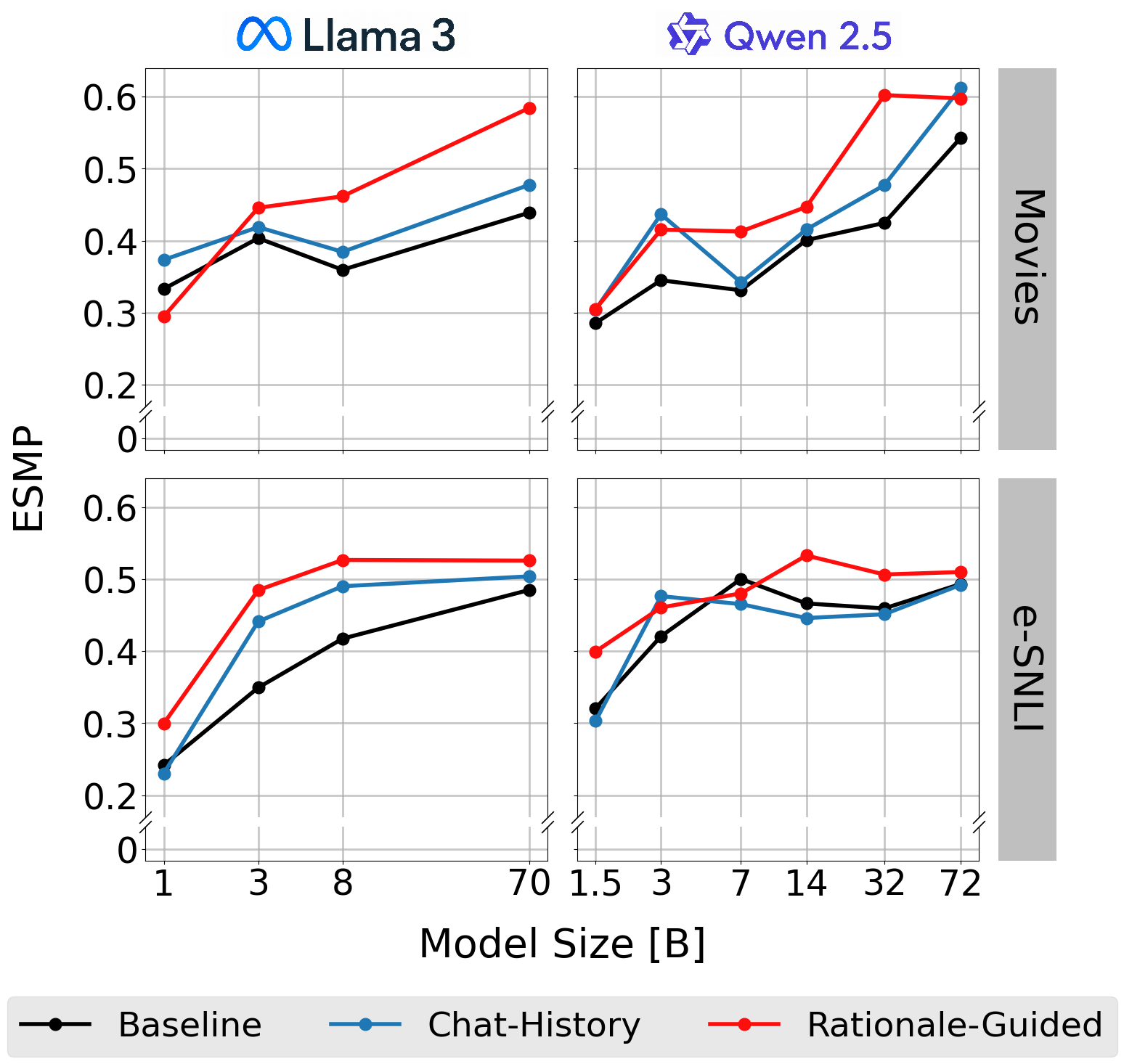}
\caption{ESMP shows that larger models produce more targeted, decision-relevant edits.}
\label{fig:esmp}
\end{figure}

\textbf{Faithful self-explanations are also more human-aligned.}
Larger models not only produce more faithful self-explanations; their self-explanations also become more aligned with human rationales (Figure~\ref{fig:esmp}). ESMP captures this aspect by measuring whether the evidence edited by the model overlaps with the evidence annotated by humans. This matters because, in our setting, a self-explanation is meaningful when the model changes its own prediction by intervening on the evidence that supports that prediction. Higher ESMP suggests that larger models explain themselves through rationales that are closer to those used by humans.

This trend is systematic: model size is strongly associated with ESMP ($\rho=0.78$), while its association with closeness ($\rho=0.32$) and semantic similarity ($\rho=0.16$) is much weaker. Thus, scaling does not mainly make self-explanations longer, broader, or more text-preserving. Instead, it makes models better at identifying the evidence behind their own decisions that is more consistent with human reasoning.

\textbf{Prompting can distort self-explanations.}
Chat-History produces the most faithful self-explanations, achieving the best score in 12 out of 20 settings. The Rationale-Guided condition, however, lowers faithfulness while improving ESMP and closeness. This happens because this setting explicitly pushes the model to identify evidence-like tokens before editing them, making the explanation look more human-aligned and minimal. Since faithfulness decreases, these edits do not better reflect what drives the model's own decision; rather, they reflect the structure imposed by the prompt. Thus, Rationale-Guided Counterfactual Generation can produce self-explanations that appear stronger under human-alignment metrics, while being less faithful to the model's actual behavior.

\textbf{Task structure matters.}
The two tasks show different patterns. In Movies, self-explanations can often be produced through local lexical substitutions, such as changing sentiment-bearing words, while preserving the overall meaning of the input. This explains why larger models can improve flip rate, ESMP, and semantic similarity at the same time (Figures \ref{fig:faithfulness}, \ref{fig:minimality}, and \ref{fig:esmp} respectively). In e-SNLI, however, changing the entailment relation often requires more semantic edits involving entities, actions, or relations. As a result, larger models improve flip rate and ESMP, but semantic similarity does not increase in the same way.

\section{Conclusions}

In this work, we examined whether LLMs can explain their own predictions through counterfactual self-explanations. Our results show that this ability depends primarily on model scale: larger models produce more faithful self-explanations, i.e., edits that change their own decisions, while increasingly targeting evidence aligned with human rationales. Thus, stronger models do not merely generate better-looking explanations, but better localize the evidence behind their predictions. At the same time, prompting can make self-explanations appear more convincing without making them more faithful. In particular, the Rationale-Guided condition produces counterfactuals that are more minimal under the normalized edit-distance metric and more human-aligned, but less faithful, suggesting that such rationales may reflect the prompt structure rather than the model's actual decision process. Overall, self-generated counterfactuals provide useful behavioral evidence of what a model relies on, but should not be treated as direct access to its internal reasoning.

\section*{Limitations}

Herein we acknowledge certain limitations of the current work. We restrict our analysis to binary classification tasks, where model behavior is expressed through discrete predictions. This enables counterfactual and rationale-based evaluation, but limits generalization to multiple-choice and open-ended generative settings. Additionally, explanations are assessed using metrics like faithfulness and minimality. Although informative, they may not fully capture human understandability or usefulness, which is usually the end goal of explainability.

\bibliography{custom}

@inproceedings{madsen2024self,
    title = "Are self-explanations from Large Language Models faithful?",
    author = "Madsen, Andreas  and
      Chandar, Sarath  and
      Reddy, Siva",
    editor = "Ku, Lun-Wei  and
      Martins, Andre  and
      Srikumar, Vivek",
    booktitle = "Findings of the Association for Computational Linguistics: ACL 2024",
    month = aug,
    year = "2024",
    address = "Bangkok, Thailand",
    publisher = "Association for Computational Linguistics",
    url = "https://aclanthology.org/2024.findings-acl.19/",
    doi = "10.18653/v1/2024.findings-acl.19",
    pages = "295--337"
}

@inproceedings{bhattacharjee2405zero,
  title={Zero-shot LLM-guided counterfactual generation: A case study on NLP model evaluation},
  author={Bhattacharjee, Amrita and Moraffah, Raha and Garland, Joshua and Liu, Huan},
  booktitle={2024 IEEE International Conference on Big Data (BigData)},
  pages={1243--1248},
  year={2024},
  organization={IEEE}
}

@inproceedings{nguyen2024llmsgeneratingevaluatingcounterfactuals,
    title = "{LLM}s for Generating and Evaluating Counterfactuals: A Comprehensive Study",
    author = {Nguyen, Van Bach  and
      Youssef, Paul  and
      Seifert, Christin  and
      Schl{\"o}tterer, J{\"o}rg},
    editor = "Al-Onaizan, Yaser  and
      Bansal, Mohit  and
      Chen, Yun-Nung",
    booktitle = "Findings of the Association for Computational Linguistics: EMNLP 2024",
    month = nov,
    year = "2024",
    address = "Miami, Florida, USA",
    publisher = "Association for Computational Linguistics",
    url = "https://aclanthology.org/2024.findings-emnlp.870/",
    doi = "10.18653/v1/2024.findings-emnlp.870",
    pages = "14809--14824"
}

@book{molnar2020interpretable,
  title={Interpretable machine learning},
  author={Molnar, Christoph},
  year={2020},
  publisher={{Lulu.com}}
}

@misc{grattafiori2024llama3herdmodels,
      title={The Llama 3 Herd of Models}, 
      author={Aaron Grattafiori and Abhimanyu Dubey and Abhinav Jauhri and Abhinav Pandey and Abhishek Kadian and Ahmad Al-Dahle and Aiesha Letman and Akhil Mathur and Alan Schelten and Alex Vaughan and Amy Yang and Angela Fan and Anirudh Goyal and Anthony Hartshorn and Aobo Yang and Archi Mitra and Archie Sravankumar and Artem Korenev and Arthur Hinsvark and Arun Rao and Aston Zhang and Aurelien Rodriguez and Austen Gregerson and Ava Spataru and Baptiste Roziere and Bethany Biron and Binh Tang and Bobbie Chern and Charlotte Caucheteux and Chaya Nayak and Chloe Bi and Chris Marra and Chris McConnell and Christian Keller and Christophe Touret and Chunyang Wu and Corinne Wong and Cristian Canton Ferrer and Cyrus Nikolaidis and Damien Allonsius and Daniel Song and Danielle Pintz and Danny Livshits and Danny Wyatt and David Esiobu and Dhruv Choudhary and Dhruv Mahajan and Diego Garcia-Olano and Diego Perino and Dieuwke Hupkes and Egor Lakomkin and Ehab AlBadawy and Elina Lobanova and Emily Dinan and Eric Michael Smith and Filip Radenovic and Francisco Guzmán and Frank Zhang and Gabriel Synnaeve and Gabrielle Lee and Georgia Lewis Anderson and Govind Thattai and Graeme Nail and Gregoire Mialon and Guan Pang and Guillem Cucurell and Hailey Nguyen and Hannah Korevaar and Hu Xu and Hugo Touvron and Iliyan Zarov and Imanol Arrieta Ibarra and Isabel Kloumann and Ishan Misra and Ivan Evtimov and Jack Zhang and Jade Copet and Jaewon Lee and Jan Geffert and Jana Vranes and Jason Park and Jay Mahadeokar and Jeet Shah and Jelmer van der Linde and Jennifer Billock and Jenny Hong and Jenya Lee and Jeremy Fu and Jianfeng Chi and Jianyu Huang and Jiawen Liu and Jie Wang and Jiecao Yu and Joanna Bitton and Joe Spisak and Jongsoo Park and Joseph Rocca and Joshua Johnstun and Joshua Saxe and Junteng Jia and Kalyan Vasuden Alwala and Karthik Prasad and Kartikeya Upasani and Kate Plawiak and Ke Li and Kenneth Heafield and Kevin Stone and Khalid El-Arini and Krithika Iyer and Kshitiz Malik and Kuenley Chiu and Kunal Bhalla and Kushal Lakhotia and Lauren Rantala-Yeary and Laurens van der Maaten and Lawrence Chen and Liang Tan and Liz Jenkins and Louis Martin and Lovish Madaan and Lubo Malo and Lukas Blecher and Lukas Landzaat and Luke de Oliveira and Madeline Muzzi and Mahesh Pasupuleti and Mannat Singh and Manohar Paluri and Marcin Kardas and Maria Tsimpoukelli and Mathew Oldham and Mathieu Rita and Maya Pavlova and Melanie Kambadur and Mike Lewis and Min Si and Mitesh Kumar Singh and Mona Hassan and Naman Goyal and Narjes Torabi and Nikolay Bashlykov and Nikolay Bogoychev and Niladri Chatterji and Ning Zhang and Olivier Duchenne and Onur Çelebi and Patrick Alrassy and Pengchuan Zhang and Pengwei Li and Petar Vasic and Peter Weng and Prajjwal Bhargava and Pratik Dubal and Praveen Krishnan and Punit Singh Koura and Puxin Xu and Qing He and Qingxiao Dong and Ragavan Srinivasan and Raj Ganapathy and Ramon Calderer and Ricardo Silveira Cabral and Robert Stojnic and Roberta Raileanu and Rohan Maheswari and Rohit Girdhar and Rohit Patel and Romain Sauvestre and Ronnie Polidoro and Roshan Sumbaly and Ross Taylor and Ruan Silva and Rui Hou and Rui Wang and Saghar Hosseini and Sahana Chennabasappa and Sanjay Singh and Sean Bell and Seohyun Sonia Kim and Sergey Edunov and Shaoliang Nie and Sharan Narang and Sharath Raparthy and Sheng Shen and Shengye Wan and Shruti Bhosale and Shun Zhang and Simon Vandenhende and Soumya Batra and Spencer Whitman and Sten Sootla and Stephane Collot and Suchin Gururangan and Sydney Borodinsky and Tamar Herman and Tara Fowler and Tarek Sheasha and Thomas Georgiou and Thomas Scialom and Tobias Speckbacher and Todor Mihaylov and Tong Xiao and Ujjwal Karn and Vedanuj Goswami and Vibhor Gupta and Vignesh Ramanathan and Viktor Kerkez and Vincent Gonguet and Virginie Do and Vish Vogeti and Vítor Albiero and Vladan Petrovic and Weiwei Chu and Wenhan Xiong and Wenyin Fu and Whitney Meers and Xavier Martinet and Xiaodong Wang and Xiaofang Wang and Xiaoqing Ellen Tan and Xide Xia and Xinfeng Xie and Xuchao Jia and Xuewei Wang and Yaelle Goldschlag and Yashesh Gaur and Yasmine Babaei and Yi Wen and Yiwen Song and Yuchen Zhang and Yue Li and Yuning Mao and Zacharie Delpierre Coudert and Zheng Yan and Zhengxing Chen and Zoe Papakipos and Aaditya Singh and Aayushi Srivastava and Abha Jain and Adam Kelsey and Adam Shajnfeld and Adithya Gangidi and Adolfo Victoria and Ahuva Goldstand and Ajay Menon and Ajay Sharma and Alex Boesenberg and Alexei Baevski and Allie Feinstein and Amanda Kallet and Amit Sangani and Amos Teo and Anam Yunus and Andrei Lupu and Andres Alvarado and Andrew Caples and Andrew Gu and Andrew Ho and Andrew Poulton and Andrew Ryan and Ankit Ramchandani and Annie Dong and Annie Franco and Anuj Goyal and Aparajita Saraf and Arkabandhu Chowdhury and Ashley Gabriel and Ashwin Bharambe and Assaf Eisenman and Azadeh Yazdan and Beau James and Ben Maurer and Benjamin Leonhardi and Bernie Huang and Beth Loyd and Beto De Paola and Bhargavi Paranjape and Bing Liu and Bo Wu and Boyu Ni and Braden Hancock and Bram Wasti and Brandon Spence and Brani Stojkovic and Brian Gamido and Britt Montalvo and Carl Parker and Carly Burton and Catalina Mejia and Ce Liu and Changhan Wang and Changkyu Kim and Chao Zhou and Chester Hu and Ching-Hsiang Chu and Chris Cai and Chris Tindal and Christoph Feichtenhofer and Cynthia Gao and Damon Civin and Dana Beaty and Daniel Kreymer and Daniel Li and David Adkins and David Xu and Davide Testuggine and Delia David and Devi Parikh and Diana Liskovich and Didem Foss and Dingkang Wang and Duc Le and Dustin Holland and Edward Dowling and Eissa Jamil and Elaine Montgomery and Eleonora Presani and Emily Hahn and Emily Wood and Eric-Tuan Le and Erik Brinkman and Esteban Arcaute and Evan Dunbar and Evan Smothers and Fei Sun and Felix Kreuk and Feng Tian and Filippos Kokkinos and Firat Ozgenel and Francesco Caggioni and Frank Kanayet and Frank Seide and Gabriela Medina Florez and Gabriella Schwarz and Gada Badeer and Georgia Swee and Gil Halpern and Grant Herman and Grigory Sizov and Guangyi and Zhang and Guna Lakshminarayanan and Hakan Inan and Hamid Shojanazeri and Han Zou and Hannah Wang and Hanwen Zha and Haroun Habeeb and Harrison Rudolph and Helen Suk and Henry Aspegren and Hunter Goldman and Hongyuan Zhan and Ibrahim Damlaj and Igor Molybog and Igor Tufanov and Ilias Leontiadis and Irina-Elena Veliche and Itai Gat and Jake Weissman and James Geboski and James Kohli and Janice Lam and Japhet Asher and Jean-Baptiste Gaya and Jeff Marcus and Jeff Tang and Jennifer Chan and Jenny Zhen and Jeremy Reizenstein and Jeremy Teboul and Jessica Zhong and Jian Jin and Jingyi Yang and Joe Cummings and Jon Carvill and Jon Shepard and Jonathan McPhie and Jonathan Torres and Josh Ginsburg and Junjie Wang and Kai Wu and Kam Hou U and Karan Saxena and Kartikay Khandelwal and Katayoun Zand and Kathy Matosich and Kaushik Veeraraghavan and Kelly Michelena and Keqian Li and Kiran Jagadeesh and Kun Huang and Kunal Chawla and Kyle Huang and Lailin Chen and Lakshya Garg and Lavender A and Leandro Silva and Lee Bell and Lei Zhang and Liangpeng Guo and Licheng Yu and Liron Moshkovich and Luca Wehrstedt and Madian Khabsa and Manav Avalani and Manish Bhatt and Martynas Mankus and Matan Hasson and Matthew Lennie and Matthias Reso and Maxim Groshev and Maxim Naumov and Maya Lathi and Meghan Keneally and Miao Liu and Michael L. Seltzer and Michal Valko and Michelle Restrepo and Mihir Patel and Mik Vyatskov and Mikayel Samvelyan and Mike Clark and Mike Macey and Mike Wang and Miquel Jubert Hermoso and Mo Metanat and Mohammad Rastegari and Munish Bansal and Nandhini Santhanam and Natascha Parks and Natasha White and Navyata Bawa and Nayan Singhal and Nick Egebo and Nicolas Usunier and Nikhil Mehta and Nikolay Pavlovich Laptev and Ning Dong and Norman Cheng and Oleg Chernoguz and Olivia Hart and Omkar Salpekar and Ozlem Kalinli and Parkin Kent and Parth Parekh and Paul Saab and Pavan Balaji and Pedro Rittner and Philip Bontrager and Pierre Roux and Piotr Dollar and Polina Zvyagina and Prashant Ratanchandani and Pritish Yuvraj and Qian Liang and Rachad Alao and Rachel Rodriguez and Rafi Ayub and Raghotham Murthy and Raghu Nayani and Rahul Mitra and Rangaprabhu Parthasarathy and Raymond Li and Rebekkah Hogan and Robin Battey and Rocky Wang and Russ Howes and Ruty Rinott and Sachin Mehta and Sachin Siby and Sai Jayesh Bondu and Samyak Datta and Sara Chugh and Sara Hunt and Sargun Dhillon and Sasha Sidorov and Satadru Pan and Saurabh Mahajan and Saurabh Verma and Seiji Yamamoto and Sharadh Ramaswamy and Shaun Lindsay and Shaun Lindsay and Sheng Feng and Shenghao Lin and Shengxin Cindy Zha and Shishir Patil and Shiva Shankar and Shuqiang Zhang and Shuqiang Zhang and Sinong Wang and Sneha Agarwal and Soji Sajuyigbe and Soumith Chintala and Stephanie Max and Stephen Chen and Steve Kehoe and Steve Satterfield and Sudarshan Govindaprasad and Sumit Gupta and Summer Deng and Sungmin Cho and Sunny Virk and Suraj Subramanian and Sy Choudhury and Sydney Goldman and Tal Remez and Tamar Glaser and Tamara Best and Thilo Koehler and Thomas Robinson and Tianhe Li and Tianjun Zhang and Tim Matthews and Timothy Chou and Tzook Shaked and Varun Vontimitta and Victoria Ajayi and Victoria Montanez and Vijai Mohan and Vinay Satish Kumar and Vishal Mangla and Vlad Ionescu and Vlad Poenaru and Vlad Tiberiu Mihailescu and Vladimir Ivanov and Wei Li and Wenchen Wang and Wenwen Jiang and Wes Bouaziz and Will Constable and Xiaocheng Tang and Xiaojian Wu and Xiaolan Wang and Xilun Wu and Xinbo Gao and Yaniv Kleinman and Yanjun Chen and Ye Hu and Ye Jia and Ye Qi and Yenda Li and Yilin Zhang and Ying Zhang and Yossi Adi and Youngjin Nam and Yu and Wang and Yu Zhao and Yuchen Hao and Yundi Qian and Yunlu Li and Yuzi He and Zach Rait and Zachary DeVito and Zef Rosnbrick and Zhaoduo Wen and Zhenyu Yang and Zhiwei Zhao and Zhiyu Ma},
      year={2024},
      eprint={2407.21783},
      archivePrefix={arXiv},
      primaryClass={cs.AI},
      url={https://arxiv.org/abs/2407.21783}, 
}

@misc{qwen25technicalreport,
      title={Qwen2.5 Technical Report}, 
      author={An Yang and Baosong Yang and Beichen Zhang and Binyuan Hui and Bo Zheng and Bowen Yu and Chengyuan Li and Dayiheng Liu and Fei Huang and Haoran Wei and Huan Lin and Jian Yang and Jianhong Tu and Jianwei Zhang and Jianxin Yang and Jiaxi Yang and Jingren Zhou and Junyang Lin and Kai Dang and Keming Lu and Keqin Bao and Kexin Yang and Le Yu and Mei Li and Mingfeng Xue and Pei Zhang and Qin Zhu and Rui Men and Runji Lin and Tianhao Li and Tianyi Tang and Tingyu Xia and Xingzhang Ren and Xuancheng Ren and Yang Fan and Yang Su and Yichang Zhang and Yu Wan and Yuqiong Liu and Zeyu Cui and Zhenru Zhang and Zihan Qiu},
      year={2025},
      eprint={2412.15115},
      archivePrefix={arXiv},
      primaryClass={cs.CL},
      url={https://arxiv.org/abs/2412.15115}, 
}

@inproceedings{dehghanighobadi2025can,
    title = "Can {LLM}s Explain Themselves Counterfactually?",
    author = "Dehghanighobadi, Zahra  and
      Fischer, Asja  and
      Zafar, Muhammad Bilal",
    editor = "Christodoulopoulos, Christos  and
      Chakraborty, Tanmoy  and
      Rose, Carolyn  and
      Peng, Violet",
    booktitle = "Proceedings of the 2025 Conference on Empirical Methods in Natural Language Processing",
    month = nov,
    year = "2025",
    address = "Suzhou, China",
    publisher = "Association for Computational Linguistics",
    url = "https://aclanthology.org/2025.emnlp-main.396/",
    doi = "10.18653/v1/2025.emnlp-main.396",
    pages = "7787--7815",
    ISBN = "979-8-89176-332-6"
}

@inproceedings{deyoung-etal-2020-eraser,
    title = "{ERASER}: {A} Benchmark to Evaluate Rationalized {NLP} Models",
    author = "DeYoung, Jay  and
      Jain, Sarthak  and
      Rajani, Nazneen Fatema  and
      Lehman, Eric  and
      Xiong, Caiming  and
      Socher, Richard  and
      Wallace, Byron C.",
    editor = "Jurafsky, Dan  and
      Chai, Joyce  and
      Schluter, Natalie  and
      Tetreault, Joel",
    booktitle = "Proceedings of the 58th Annual Meeting of the Association for Computational Linguistics",
    month = jul,
    year = "2020",
    address = "Online",
    publisher = "Association for Computational Linguistics",
    url = "https://aclanthology.org/2020.acl-main.408/",
    doi = "10.18653/v1/2020.acl-main.408",
    pages = "4443--4458"
}

@inproceedings{zaidan-eisner-2008-modeling,
    title = "Modeling Annotators: {A} Generative Approach to Learning from Annotator Rationales",
    author = "Zaidan, Omar  and
      Eisner, Jason",
    editor = "Lapata, Mirella  and
      Ng, Hwee Tou",
    booktitle = "Proceedings of the 2008 Conference on Empirical Methods in Natural Language Processing",
    month = oct,
    year = "2008",
    address = "Honolulu, Hawaii",
    publisher = "Association for Computational Linguistics",
    url = "https://aclanthology.org/D08-1004/",
    pages = "31--40"
}

@inproceedings{NEURIPS2018_4c7a167b,
 author = {Camburu, Oana-Maria and Rockt\"{a}schel, Tim and Lukasiewicz, Thomas and Blunsom, Phil},
 booktitle = {Advances in Neural Information Processing Systems},
 editor = {S. Bengio and H. Wallach and H. Larochelle and K. Grauman and N. Cesa-Bianchi and R. Garnett},
 pages = {},
 publisher = {Curran Associates, Inc.},
 title = {e-SNLI: Natural Language Inference with Natural Language Explanations},
 url = {https://proceedings.neurips.cc/paper_files/paper/2018/file/4c7a167bb329bd92580a99ce422d6fa6-Paper.pdf},
 volume = {31},
 year = {2018}
}

@misc{hong2026llm,
      title={Do LLM Self-Explanations Help Users Predict Model Behavior? Evaluating Counterfactual Simulatability with Pragmatic Perturbations}, 
      author={Pingjun Hong and Benjamin Roth},
      year={2026},
      eprint={2601.03775},
      archivePrefix={arXiv},
      primaryClass={cs.CL},
      url={https://arxiv.org/abs/2601.03775}, 
}

@inproceedings{hanrfeval,
  title={RFEval: Benchmarking Reasoning Faithfulness under Counterfactual Reasoning Intervention in Large Reasoning Models},
  author={Han, Yunseok and Lee, Yejoon and Do, Jaeyoung},
  year={2026},
  booktitle={ICLR 2026}
}

@misc{siegel2025verbosity,
      title={Verbosity Tradeoffs and the Impact of Scale on the Faithfulness of LLM Self-Explanations}, 
      author={Noah Y. Siegel and Nicolas Heess and Maria Perez-Ortiz and Oana-Maria Camburu},
      year={2025},
      eprint={2503.13445},
      archivePrefix={arXiv},
      primaryClass={cs.CL},
      url={https://arxiv.org/abs/2503.13445}, 
}

@inproceedings{chuang2026faithlm,
    title = "{F}aith{LM}: Towards Faithful Explanations for Large Language Models",
    author = "Chuang, Yu-Neng  and
      Wang, Guanchu  and
      Chang, Chia-Yuan  and
      Tang, Ruixiang  and
      Zhong, Shaochen  and
      Yang, Fan  and
      Wen, Andrew  and
      Du, Mengnan  and
      Cai, Xuanting  and
      Braverman, Vladimir  and
      Hu, Xia",
    editor = "Demberg, Vera  and
      Inui, Kentaro  and
      Marquez, Llu{\'i}s",
    booktitle = "Proceedings of the 19th Conference of the {E}uropean Chapter of the {A}ssociation for {C}omputational {L}inguistics (Volume 1: Long Papers)",
    month = mar,
    year = "2026",
    address = "Rabat, Morocco",
    publisher = "Association for Computational Linguistics",
    url = "https://aclanthology.org/2026.eacl-long.177/",
    doi = "10.18653/v1/2026.eacl-long.177",
    pages = "3802--3824",
    ISBN = "979-8-89176-380-7"
}

@inproceedings{10.5555/3666122.3669397,
author = {Turpin, Miles and Michael, Julian and Perez, Ethan and Bowman, Samuel R.},
title = {Language models don't always say what they think: unfaithful explanations in chain-of-thought prompting},
year = {2023},
publisher = {Curran Associates Inc.},
address = {Red Hook, NY, USA},
booktitle = {Proceedings of the 37th International Conference on Neural Information Processing Systems},
articleno = {3275},
numpages = {14},
location = {New Orleans, LA, USA},
series = {NIPS '23}
}

@article{chen2025reasoning,
  title={Reasoning models don't always say what they think},
  author={Chen, Yanda and Benton, Joe and Radhakrishnan, Ansh and Uesato, Jonathan and Denison, Carson and Schulman, John and Somani, Arushi and Hase, Peter and Wagner, Misha and Roger, Fabien and others},
  journal={arXiv preprint arXiv:2505.05410},
  year={2025}
}

@inproceedings{atanasova-etal-2023-faithfulness,
    title = "Faithfulness Tests for Natural Language Explanations",
    author = "Atanasova, Pepa  and
      Camburu, Oana-Maria  and
      Lioma, Christina  and
      Lukasiewicz, Thomas  and
      Simonsen, Jakob Grue  and
      Augenstein, Isabelle",
    editor = "Rogers, Anna  and
      Boyd-Graber, Jordan  and
      Okazaki, Naoaki",
    booktitle = "Proceedings of the 61st Annual Meeting of the Association for Computational Linguistics (Volume 2: Short Papers)",
    month = jul,
    year = "2023",
    address = "Toronto, Canada",
    publisher = "Association for Computational Linguistics",
    url = "https://aclanthology.org/2023.acl-short.25/",
    doi = "10.18653/v1/2023.acl-short.25",
    pages = "283--294"
}

@inproceedings{jacovi-goldberg-2020-towards,
    title = "Towards Faithfully Interpretable {NLP} Systems: How Should We Define and Evaluate Faithfulness?",
    author = "Jacovi, Alon  and
      Goldberg, Yoav",
    editor = "Jurafsky, Dan  and
      Chai, Joyce  and
      Schluter, Natalie  and
      Tetreault, Joel",
    booktitle = "Proceedings of the 58th Annual Meeting of the Association for Computational Linguistics",
    month = jul,
    year = "2020",
    address = "Online",
    publisher = "Association for Computational Linguistics",
    url = "https://aclanthology.org/2020.acl-main.386/",
    doi = "10.18653/v1/2020.acl-main.386",
    pages = "4198--4205"
}

@inproceedings{siegel-etal-2024-probabilities,
    title = "The Probabilities Also Matter: A More Faithful Metric for Faithfulness of Free-Text Explanations in Large Language Models",
    author = "Siegel, Noah  and
      Camburu, Oana-Maria  and
      Heess, Nicolas  and
      Perez-Ortiz, Maria",
    editor = "Ku, Lun-Wei  and
      Martins, Andre  and
      Srikumar, Vivek",
    booktitle = "Proceedings of the 62nd Annual Meeting of the Association for Computational Linguistics (Volume 2: Short Papers)",
    month = aug,
    year = "2024",
    address = "Bangkok, Thailand",
    publisher = "Association for Computational Linguistics",
    url = "https://aclanthology.org/2024.acl-short.49/",
    doi = "10.18653/v1/2024.acl-short.49",
    pages = "530--546"
}

@article{randl2025mind,
  author  = {Randl, Korbinian and Pavlopoulos, John and Henriksson, Aron and Lindgren, Tony},
  title   = {Mind the gap: from plausible to valid self-explanations in large language models},
  journal = {Machine Learning},
  volume  = {114},
  year    = {2025},
  article = {220},
  doi     = {10.1007/s10994-025-06838-6}
}

@inproceedings{mayne-etal-2025-llms,
    title = "{LLM}s Don{'}t Know Their Own Decision Boundaries: The Unreliability of Self-Generated Counterfactual Explanations",
    author = "Mayne, Harry  and
      Kearns, Ryan Othniel  and
      Yang, Yushi  and
      Bean, Andrew M.  and
      Delaney, Eoin D.  and
      Russell, Chris  and
      Mahdi, Adam",
    editor = "Christodoulopoulos, Christos  and
      Chakraborty, Tanmoy  and
      Rose, Carolyn  and
      Peng, Violet",
    booktitle = "Proceedings of the 2025 Conference on Empirical Methods in Natural Language Processing",
    month = nov,
    year = "2025",
    address = "Suzhou, China",
    publisher = "Association for Computational Linguistics",
    url = "https://aclanthology.org/2025.emnlp-main.1231/",
    doi = "10.18653/v1/2025.emnlp-main.1231",
    pages = "24161--24186",
    ISBN = "979-8-89176-332-6"
}

@inproceedings{filandrianos-etal-2023-counterfactuals,
    title = "Counterfactuals of Counterfactuals: a back-translation-inspired approach to analyse counterfactual editors",
    author = "Filandrianos, George  and
      Dervakos, Edmund  and
      Menis Mastromichalakis, Orfeas  and
      Zerva, Chrysoula  and
      Stamou, Giorgos",
    editor = "Rogers, Anna  and
      Boyd-Graber, Jordan  and
      Okazaki, Naoaki",
    booktitle = "Findings of the Association for Computational Linguistics: ACL 2023",
    month = jul,
    year = "2023",
    address = "Toronto, Canada",
    publisher = "Association for Computational Linguistics",
    url = "https://aclanthology.org/2023.findings-acl.606/",
    doi = "10.18653/v1/2023.findings-acl.606",
    pages = "9507--9525"
}

@inproceedings{mastromichalakis2025beyond,
  title     = {Beyond One-Size-Fits-All: How User Objectives Shape Counterfactual Explanations},
  author    = {Menis Mastromichalakis, Orfeas and Liartis, Jason and Stamou, Giorgos},
  booktitle = {Proceedings of the XAI 2025 Late-breaking Work, Demos and Doctoral Consortium of the 3rd World Conference on eXplainable Artificial Intelligence (XAI 2025)},
  year      = {2025},
  url       = {https://ceur-ws.org/Vol-4017/paper_15.pdf}
}

\appendix

\section{Model Identifiers}
\label{app:model-ids}

Table~\ref{tab:model-ids} lists the exact model identifiers used in the experiments. 
All models were executed locally on a dedicated GPU-based infrastructure, rather than through external APIs or hosted inference services. 
This setup allowed us to maintain full control over the inference environment and ensured that all models were evaluated under comparable computational conditions. 
The experiments were conducted using a single NVIDIA H200 GPU with 140 GB of available GPU memory.
The decoding configuration used during inference was defined by the following parameters: temperature [0], top-$p$ [1], top-$k$ [0], maximum new tokens [4096], and sampling [disabled].

\begin{table}[!h]
\centering
\small
\begin{tabular}{ll}
\toprule
Family & Model Identifier \\
\midrule
Qwen & \texttt{Qwen/Qwen2.5-1.5B-Instruct} \\
Qwen & \texttt{Qwen/Qwen2.5-3B-Instruct} \\
Qwen & \texttt{Qwen/Qwen2.5-7B-Instruct} \\
Qwen & \texttt{Qwen/Qwen2.5-14B-Instruct} \\
Qwen & \texttt{Qwen/Qwen2.5-32B-Instruct} \\
Qwen & \texttt{Qwen/Qwen2.5-72B-Instruct} \\
\midrule
LLaMA & \texttt{meta-llama/Llama-3.2-1B-Instruct} \\
LLaMA & \texttt{meta-llama/Llama-3.2-3B-Instruct} \\
LLaMA & \texttt{meta-llama/Llama-3.1-8B-Instruct} \\
LLaMA & \texttt{meta-llama/Llama-3.1-70B-Instruct} \\
\midrule
MPNet & \texttt{sentence-transformers/all-mpnet-base-v2} \\
\bottomrule
\end{tabular}
\caption{Exact model identifiers used in the experiments.}
\label{tab:model-ids}
\end{table}

\section{Prompts}
\label{app:prompts}

The prompts used for the two tasks under the baseline setup are presented below.

\begin{tcolorbox}[
    colback=gray!3!white,
    colframe=black!65,
    title=\textbf{Movies task - Baseline},
    fonttitle=\bfseries,
    sharp corners=south,
    boxrule=0.6pt,
    arc=2mm,
    left=1.5mm,
    right=1.5mm,
    top=1mm,
    bottom=1mm,
    breakable
]
\small

\textbf{Stage 1: Original Classification}

\vspace{0.3em}
\textbf{System Instruction.}  
You are a sentiment classifier. Answer only ``positive'' or ``negative''. Do not explain the answer. What is the sentiment of the user's paragraph?

\vspace{0.3em}
\textbf{User.}  
Paragraph: The movie was dull, predictable, and failed to engage the audience.

\vspace{0.3em}
\textbf{Model.}  
\emph{Negative.}

\vspace{0.8em}
\hrule
\vspace{0.8em}

\textbf{Stage 2: Counterfactual Generation}

\vspace{0.3em}
\textbf{System Instruction.}  
The task is sentiment classification. The following paragraph was classified as ``negative''. Generate a counterfactual explanation by making minimal changes to the paragraph, so that the classifier would predict that the paragraph is ``positive''. Use the following definition of `counterfactual explanation': ``A counterfactual explanation is a minimal edit of the original paragraph with the words or phrases crucial for classification changed, revealing what should have been different to observe the opposite outcome.'' Enclose only the edited paragraph inside \texttt{<new>} tags and nothing else.

\vspace{0.3em}
\textbf{User.}  
The movie was dull, predictable, and failed to engage the audience.

\vspace{0.3em}
\textbf{Model.}  
\emph{\texttt{<new>}The movie was engaging, enjoyable, and pleasantly surprising.\texttt{</new>}}

\vspace{0.8em}
\hrule
\vspace{0.8em}

\textbf{Stage 3: Counterfactual Classification}

\vspace{0.3em}
\textbf{System Instruction.}  
You are a sentiment classifier. Answer only ``positive'' or ``negative''. Do not explain the answer. What is the sentiment of the user's paragraph?

\vspace{0.3em}
\textbf{User.}  
Paragraph: The movie was engaging, enjoyable, and pleasantly surprising.

\vspace{0.3em}
\textbf{Model.}  
\emph{Positive.}

\end{tcolorbox}

\begin{tcolorbox}[
    colback=gray!3!white,
    colframe=black!65,
    title=\textbf{SNLI task - Baseline},
    fonttitle=\bfseries,
    sharp corners=south,
    boxrule=0.6pt,
    arc=2mm,
    left=1.5mm,
    right=1.5mm,
    top=1mm,
    bottom=1mm,
    breakable
]
\small

\textbf{Stage 1: Original Classification}

\vspace{0.3em}
\textbf{System Instruction.}  
You are an entailment classifier. Does the statement ``A dog is chasing a ball across a grassy field.'' entail from the following paragraph? Answer either ``yes'' for entailment or ``no'' for no entailment. Do not explain the answer.

\vspace{0.3em}
\textbf{User.}  
Paragraph: An animal is running outside.

\vspace{0.3em}
\textbf{Model.}  
\emph{Yes.}

\vspace{0.8em}
\hrule
\vspace{0.8em}

\textbf{Stage 2: Counterfactual Generation}

\vspace{0.3em}
\textbf{System Instruction.}  
The task is entailment classification and the statement ``A dog is chasing a ball across a grassy field.'' entails from the user's paragraph. Generate a counterfactual explanation by making minimal changes to the paragraph, so that the following paragraph does not entail the statement ``A dog is chasing a ball across a grassy field.''. Use the following definition of `counterfactual explanation': ``A counterfactual explanation is a minimal edit of the original paragraph with the words or phrases crucial for classification changed, revealing what should have been different to observe the opposite outcome.'' Enclose only the edited paragraph inside \texttt{<new>} tags and nothing else; for example: \texttt{<new>}The revised paragraph goes here.\texttt{</new>}.

\vspace{0.3em}
\textbf{User.}  
Paragraph: An animal is running outside.

\vspace{0.3em}
\textbf{Model.}  
\emph{\texttt{<new>}A cat is running outside.\texttt{</new>}}

\vspace{0.8em}
\hrule
\vspace{0.8em}

\textbf{Stage 3: Counterfactual Classification}

\vspace{0.3em}
\textbf{System Instruction.}  
You are an entailment classifier. Does the statement ``A dog is chasing a ball across a grassy field.'' entail from the following paragraph? Answer either ``yes'' for entailment or ``no'' for no entailment. Do not explain the answer.

\vspace{0.3em}
\textbf{User.}  
Paragraph: A cat is running outside.

\vspace{0.3em}
\textbf{Model.}  
\emph{No.}

\end{tcolorbox}

The prompts used for the two tasks under the Rationale-Guided condition are presented below.
\begin{tcolorbox}[
    colback=gray!3!white,
    colframe=black!65,
    title=\textbf{Movies task - Rationale Guided},
    fonttitle=\bfseries,
    sharp corners=south,
    boxrule=0.6pt,
    arc=2mm,
    left=1.5mm,
    right=1.5mm,
    top=1mm,
    bottom=1mm,
    breakable
]
\small

\textbf{Stage 1: Original Classification}

\vspace{0.3em}
\textbf{System Instruction.}  
You are a sentiment classifier. Answer only ``positive'' or ``negative''. Do not explain the answer. What is the sentiment of the user's paragraph?

\vspace{0.3em}
\textbf{User.}  
Paragraph: The movie was dull, predictable, and failed to engage the audience.

\vspace{0.3em}
\textbf{Model.}  
\emph{Negative.}

\vspace{0.8em}
\hrule
\vspace{0.8em}

\textbf{Stage 2: Counterfactual Generation}

\vspace{0.3em}
\textbf{System Instruction.}  
In the task of sentiment classification, the following paragraph was classified as ``negative''. Explain why the ``negative'' label was predicted by identifying the words in the input that caused the label. List ONLY the words as a comma separated list.

\vspace{0.3em}
\textbf{User.}  
The movie was dull, predictable, and failed to engage the audience.

\vspace{0.3em}
\textbf{Model.}  
\emph{dull, predictable, failed, engage}

\vspace{0.6em}
\textbf{System Instruction.}  
Generate a counterfactual explanation for the original paragraph by ONLY changing a minimal set of the words you identified, so that the classifier would predict that the paragraph is ``positive''. Use the following definition of `counterfactual explanation': ``A counterfactual explanation is a minimal edit of the original paragraph with the words or phrases crucial for classification changed, revealing what should have been different to observe the opposite outcome.'' Enclose only the edited paragraph inside \texttt{<new>} tags and nothing else.

\vspace{0.3em}
\textbf{User.}  
The movie was dull, predictable, and failed to engage the audience.

\vspace{0.3em}
\textbf{Model.}  
\emph{\texttt{<new>}The movie was engaging, enjoyable, and managed to engage the audience.\texttt{</new>}}

\vspace{0.8em}
\hrule
\vspace{0.8em}

\textbf{Stage 3: Counterfactual Classification}

\vspace{0.3em}
\textbf{System Instruction.}  
You are a sentiment classifier. Answer only ``positive'' or ``negative''. Do not explain the answer. What is the sentiment of the user's paragraph?

\vspace{0.3em}
\textbf{User.}  
Paragraph: The movie was engaging, enjoyable, and managed to engage the audience.

\vspace{0.3em}
\textbf{Model.}  
\emph{Positive.}

\end{tcolorbox}

\begin{tcolorbox}[
    colback=gray!3!white,
    colframe=black!65,
    title=\textbf{SNLI task - Rationale Guided},
    fonttitle=\bfseries,
    sharp corners=south,
    boxrule=0.6pt,
    arc=2mm,
    left=1.5mm,
    right=1.5mm,
    top=1mm,
    bottom=1mm,
    breakable
]
\small

\textbf{Stage 1: Original Classification}

\vspace{0.3em}
\textbf{System Instruction.}  
You are an entailment classifier. Does the statement ``A dog is chasing a ball across a grassy field.'' entail from the following paragraph? Answer either ``yes'' for entailment or ``no'' for no entailment. Do not explain the answer.

\vspace{0.3em}
\textbf{User.}  
Paragraph: An animal is running outside.

\vspace{0.3em}
\textbf{Model.}  
\emph{Yes.}

\vspace{0.8em}
\hrule
\vspace{0.8em}

\textbf{Stage 2: Counterfactual Generation}

\vspace{0.3em}
\textbf{System Instruction.}  
In the task of entailment classification the statement ``A dog is chasing a ball across a grassy field.'' entails from the user's paragraph. Explain the prediction by identifying the words in the paragraph that caused this label. List ONLY the words as a comma separated list.

\vspace{0.3em}
\textbf{User.}  
Paragraph: An animal is running outside.

\vspace{0.3em}
\textbf{Model.}  
\emph{animal, running, outside}

\vspace{0.6em}
\textbf{System Instruction.}  
Generate a counterfactual explanation for the original paragraph by ONLY changing a minimal set of the words you identified, so that the following paragraph does not entail the statement ``A dog is chasing a ball across a grassy field.''. Use the following definition of `counterfactual explanation': ``A counterfactual explanation is a minimal edit of the original paragraph with the words or phrases crucial for classification changed, revealing what should have been different to observe the opposite outcome.'' Enclose only the edited paragraph inside \texttt{<new>} tags and nothing else; for example: \texttt{<new>}The revised paragraph goes here.\texttt{</new>}.

\vspace{0.3em}
\textbf{User.}  
Paragraph: An animal is running outside.

\vspace{0.3em}
\textbf{Model.}  
\emph{\texttt{<new>}A cat is running outside.\texttt{</new>}}

\vspace{0.8em}
\hrule
\vspace{0.8em}

\textbf{Stage 3: Counterfactual Classification}

\vspace{0.3em}
\textbf{System Instruction.}  
You are an entailment classifier. Does the statement ``A dog is chasing a ball across a grassy field.'' entail from the following paragraph? Answer either ``yes'' for entailment or ``no'' for no entailment. Do not explain the answer.

\vspace{0.3em}
\textbf{User.}  
Paragraph: A cat is running outside.

\vspace{0.3em}
\textbf{Model.}  
\emph{No.}

\end{tcolorbox}

\section{Detailed Analysis}
\label{app:detailed-results}

Tables~\ref{tab:flip-rate}, \ref{tab:semantic-similarity}, \ref{tab:closeness}, and \ref{tab:esmp} report the full results for the two datasets across model families, model sizes, and prompting strategies, covering the four evaluation dimensions used in our analysis: faithfulness, semantic similarity, closeness, and ESMP. Hist. denotes the Chat-History setting, while RG denotes Rationale-Guided Counterfactual Generation.

\begin{table}[h]
\centering
\small
\setlength{\tabcolsep}{4pt}

\begin{tabular}{c|cc|ccc}
\hline
Dataset & Model & Size & Baseline & Hist. & RG \\
\hline

\multirow{10}{*}{\rotatebox{90}{Movies}}
& \multirow{4}{*}{LLaMA} & 1B  & 0.030 & 0.060 & 0.035 \\
&                         & 3B  & 0.437 & 0.477 & 0.462 \\
&                         & 8B  & 0.799 & 0.965 & 0.935 \\
&                         & 70B & 0.824 & 0.960 & 0.894 \\
\cline{2-6}
& \multirow{6}{*}{Qwen}  & 1.5B & 0.166 & 0.090 & 0.116 \\
&                        & 3B   & 0.216 & 0.281 & 0.286 \\
&                        & 7B   & 0.648 & 0.754 & 0.809 \\
&                        & 14B  & 0.809 & 0.859 & 0.739 \\
&                        & 32B  & 0.920 & 0.935 & 0.824 \\
&                        & 72B  & 0.940 & 0.849 & 0.829 \\
\hline

\multirow{10}{*}{\rotatebox{90}{e-SNLI}}
& \multirow{4}{*}{LLaMA} & 1B  & 0.224 & 0.261 & 0.108 \\
&                         & 3B  & 0.207 & 0.248 & 0.248 \\
&                         & 8B  & 0.648 & 0.612 & 0.643 \\
&                         & 70B & 0.872 & 0.909 & 0.873 \\
\cline{2-6}
& \multirow{6}{*}{Qwen}  & 1.5B & 0.528 & 0.460 & 0.554 \\
&                        & 3B   & 0.554 & 0.654 & 0.719 \\
&                        & 7B   & 0.800 & 0.871 & 0.854 \\
&                        & 14B  & 0.844 & 0.917 & 0.838 \\
&                        & 32B  & 0.947 & 0.961 & 0.903 \\
&                        & 72B  & 0.937 & 0.952 & 0.920 \\
\hline
\end{tabular}

\caption{\textbf{Faithfulness} results across datasets, model families, and prompting conditions. Higher values indicate that the generated self-explanation more often changes the model's original prediction.}
\label{tab:flip-rate}
\end{table}
\begin{table}[h]
\centering
\small
\setlength{\tabcolsep}{4pt}

\begin{tabular}{c|cc|ccc}
\hline
Dataset & Model & Size & Baseline & Hist. & RG \\
\hline

\multirow{10}{*}{\rotatebox{90}{Movies}}
& \multirow{4}{*}{LLaMA} & 1B  & 0.607 & 0.802 & 0.879 \\
&                         & 3B  & 0.852 & 0.814 & 0.728 \\
&                         & 8B  & 0.930 & 0.943 & 0.964 \\
&                         & 70B & 0.974 & 0.974 & 0.982 \\
\cline{2-6}
& \multirow{6}{*}{Qwen}  & 1.5B & 0.860 & 0.921 & 0.926 \\
&                        & 3B   & 0.917 & 0.943 & 0.736 \\
&                        & 7B   & 0.925 & 0.802 & 0.922 \\
&                        & 14B  & 0.966 & 0.924 & 0.924 \\
&                        & 32B  & 0.944 & 0.971 & 0.982 \\
&                        & 72B  & 0.965 & 0.981 & 0.979 \\
\hline

\multirow{10}{*}{\rotatebox{90}{e-SNLI}}
& \multirow{4}{*}{LLaMA} & 1B  & 0.784 & 0.764 & 0.806 \\
&                         & 3B  & 0.876 & 0.848 & 0.840 \\
&                         & 8B  & 0.766 & 0.763 & 0.721 \\
&                         & 70B & 0.735 & 0.716 & 0.695 \\
\cline{2-6}
& \multirow{6}{*}{Qwen}  & 1.5B & 0.726 & 0.692 & 0.704 \\
&                        & 3B   & 0.692 & 0.740 & 0.661 \\
&                        & 7B   & 0.750 & 0.701 & 0.679 \\
&                        & 14B  & 0.725 & 0.689 & 0.715 \\
&                        & 32B  & 0.700 & 0.665 & 0.698 \\
&                        & 72B  & 0.715 & 0.701 & 0.692 \\
\hline
\end{tabular}

\caption{\textbf{Semantic Similarity} results across datasets, model families, and prompting conditions. Higher values indicate stronger semantic preservation between the original input and the generated self-explanation.}
\label{tab:semantic-similarity}
\end{table}
\begin{table}[!h]
\centering
\small
\setlength{\tabcolsep}{4pt}

\begin{tabular}{c|cc|ccc}
\hline
Dataset & Model & Size & Baseline & Hist. & RG \\
\hline

\multirow{10}{*}{\rotatebox{90}{Movies}}
& \multirow{4}{*}{LLaMA} & 1B  & 0.504 & 0.756 & 0.826 \\
&                         & 3B  & 0.752 & 0.770 & 0.689 \\
&                         & 8B  & 0.700 & 0.732 & 0.864 \\
&                         & 70B & 0.883 & 0.860 & 0.916 \\
\cline{2-6}
& \multirow{6}{*}{Qwen}  & 1.5B & 0.609 & 0.754 & 0.791 \\
&                        & 3B   & 0.802 & 0.869 & 0.675 \\
&                        & 7B   & 0.728 & 0.494 & 0.770 \\
&                        & 14B  & 0.805 & 0.771 & 0.816 \\
&                        & 32B  & 0.742 & 0.827 & 0.897 \\
&                        & 72B  & 0.780 & 0.868 & 0.887 \\
\hline

\multirow{10}{*}{\rotatebox{90}{e-SNLI}}
& \multirow{4}{*}{LLaMA} & 1B  & 0.745 & 0.727 & 0.780 \\
&                         & 3B  & 0.833 & 0.843 & 0.886 \\
&                         & 8B  & 0.760 & 0.786 & 0.802 \\
&                         & 70B & 0.781 & 0.766 & 0.795 \\
\cline{2-6}
& \multirow{6}{*}{Qwen}  & 1.5B & 0.758 & 0.682 & 0.747 \\
&                        & 3B   & 0.740 & 0.796 & 0.733 \\
&                        & 7B   & 0.786 & 0.734 & 0.729 \\
&                        & 14B  & 0.753 & 0.714 & 0.795 \\
&                        & 32B  & 0.731 & 0.696 & 0.768 \\
&                        & 72B  & 0.769 & 0.748 & 0.777 \\
\hline
\end{tabular}

\caption{\textbf{Closeness} results across datasets, model families, and prompting conditions. Higher values indicate that the generated self-explanation remains closer to the original input.}
\label{tab:closeness}
\end{table}
\begin{table}[h]
\centering
\small
\setlength{\tabcolsep}{4pt}

\begin{tabular}{c|cc|ccc}
\hline
Dataset & Model & Size & Baseline & Hist. & RG \\
\hline

\multirow{10}{*}{\rotatebox{90}{Movies}}
& \multirow{4}{*}{LLaMA} & 1B  & 0.333 & 0.373 & 0.295 \\
&                         & 3B  & 0.403 & 0.419 & 0.446 \\
&                         & 8B  & 0.359 & 0.384 & 0.462 \\
&                         & 70B & 0.439 & 0.477 & 0.584 \\
\cline{2-6}
& \multirow{6}{*}{Qwen}  & 1.5B & 0.285 & 0.304 & 0.305 \\
&                        & 3B   & 0.345 & 0.437 & 0.415 \\
&                        & 7B   & 0.331 & 0.342 & 0.413 \\
&                        & 14B  & 0.401 & 0.415 & 0.447 \\
&                        & 32B  & 0.425 & 0.477 & 0.602 \\
&                        & 72B  & 0.543 & 0.613 & 0.598 \\
\hline

\multirow{10}{*}{\rotatebox{90}{e-SNLI}}
& \multirow{4}{*}{LLaMA} & 1B  & 0.242 & 0.230 & 0.300 \\
&                         & 3B  & 0.349 & 0.441 & 0.485 \\
&                         & 8B  & 0.417 & 0.490 & 0.527 \\
&                         & 70B & 0.485 & 0.504 & 0.526 \\
\cline{2-6}
& \multirow{6}{*}{Qwen}  & 1.5B & 0.321 & 0.303 & 0.400 \\
&                        & 3B   & 0.421 & 0.477 & 0.461 \\
&                        & 7B   & 0.500 & 0.465 & 0.480 \\
&                        & 14B  & 0.466 & 0.446 & 0.533 \\
&                        & 32B  & 0.460 & 0.451 & 0.507 \\
&                        & 72B  & 0.493 & 0.492 & 0.510 \\
\hline
\end{tabular}

\caption{\textbf{ESMP} results across datasets, model families, and prompting conditions. Higher values indicate stronger alignment between the model-edited evidence and human-annotated rationales.}
\label{tab:esmp}
\end{table}

In addition to the full per-model results, we performed aggregate analyses to summarize the main trends across datasets, model families, sizes, and prompting strategies. These analyses are intended to complement the tables, rather than replace the per-setting results.

First, we computed Spearman rank correlations between model size and each evaluation metric. Spearman correlation measures whether larger models tend to obtain higher values for a given metric, without assuming a linear relationship between size and performance. We find a strong positive association between model size and self-explanation faithfulness ($\rho=0.87$), and also between model size and ESMP ($\rho=0.78$). In contrast, the association is weaker for closeness ($\rho=0.32$) and much smaller for semantic similarity ($\rho=0.16$). This indicates that scaling mainly affects whether models can produce faithful and evidence-aligned self-explanations, rather than simply making the generated edits more similar to the original input.

We also fitted a regression model to estimate the effect of model size while controlling for dataset, model family, and prompting strategy. Model size was represented on a log scale, so that the coefficient can be interpreted in terms of size doublings. Under this analysis, each doubling in model size is associated with an increase of approximately 12 percentage points in self-explanation faithfulness. A complementary fractional-logit model gives the same qualitative result: each doubling in size approximately doubles the odds of producing a faithful self-explanation. These results show that the effect of scale is systematic across settings, rather than being driven by a small number of individual models.

We further compared the smallest and largest models within each family and dataset, averaging over prompting strategies. For Movies, LLaMA increases in faithfulness from 0.042 to 0.893, while Qwen increases from 0.124 to 0.873. For e-SNLI, LLaMA increases from 0.198 to 0.885, and Qwen from 0.514 to 0.936. ESMP follows the same general direction: for Movies, it increases from 0.334 to 0.500 for LLaMA and from 0.298 to 0.585 for Qwen; for e-SNLI, it increases from 0.257 to 0.505 for LLaMA and from 0.341 to 0.498 for Qwen. These comparisons show that scale improves both behavioral faithfulness and alignment with human rationales.

\begin{table*}[!h]
\centering
\small
\setlength{\tabcolsep}{4pt}
\begin{tabular}{lccccc}
\hline
Metric & Spearman $\rho$ & Spearman $p$ & $\beta$ & 95\% CI & Regression $p$ \\
\hline
Faithfulness & 0.866 & $<0.001$ & 0.119 & [0.092, 0.147] & $<0.001$ \\
ESMP & 0.778 & $<0.001$ & 0.032 & [0.023, 0.042] & $<0.001$ \\
Closeness & 0.319 & 0.013 & 0.014 & [0.005, 0.023] & 0.003 \\
Semantic Similarity & 0.157 & 0.230 & 0.010 & [-0.004, 0.023] & 0.165 \\
\hline
\end{tabular}
\caption{\textbf{Aggregate association between model size and evaluation metrics.} Spearman $\rho$ measures the monotonic association between model size and each metric. The regression coefficient $\beta$ is estimated with controls for dataset, model family, and prompting strategy, using $\log_2(\mathrm{size})$; therefore, $\beta$ corresponds to the effect of doubling model size.}
\label{tab:aggregate-size-analysis}
\end{table*}

Finally, we examined which prompting strategy performs best in each dataset--family--size setting. Chat-History achieves the highest faithfulness in 12 out of 20 settings, suggesting that preserving the original prediction in the dialogue context often helps the model generate edits that change its own subsequent decision. Rationale-Guided Counterfactual Generation achieves the highest ESMP in 15 out of 20 settings and the highest closeness in 15 out of 20 settings. This confirms that this condition tends to produce more human-like and minimal edits, even though these edits are not always the most faithful explanations of the model's own behavior.

\section{Aggregate Statistical Analysis}
\label{app:aggregate-analysis}

In addition to the full per-setting results, we report a small set of aggregate analyses to summarize the main trends across datasets, model families, model sizes, and prompting strategies. Since the same datasets, families, and prompting settings are reused across conditions, these analyses should be interpreted as descriptive evidence over aggregate metric values, rather than as instance-level causal estimates.

Table~\ref{tab:aggregate-size-analysis} shows that model size is most strongly related to faithfulness and ESMP. In the regression analysis, each doubling in model size is associated with an increase of 11.9 percentage points in faithfulness and 3.2 points in ESMP. The effects on closeness and semantic similarity are much smaller, suggesting that scale mainly improves faithful and human-aligned self-explanation, rather than simply increasing textual preservation. A complementary fractional-logit model for faithfulness gives the same qualitative conclusion: each doubling in size approximately doubles the odds of producing a faithful self-explanation (odds ratio $=2.00$, 95\% CI $[1.68, 2.40]$, $p<0.001$).

Table~\ref{tab:scale-comparison} gives a direct view of the scale effect. In all dataset--family combinations, the largest model is substantially more faithful than the smallest one, and ESMP increases in the same direction. This supports the interpretation that scale improves not only the ability to change the model's own prediction, but also the ability to do so through evidence closer to human rationales. For prompting, Chat-History achieves the highest faithfulness in 12 out of 20 settings, while Rationale-Guided achieves the highest ESMP and closeness in 15 out of 20 settings each. This supports the main observation that Chat-History is more effective for faithful self-explanation, whereas Rationale-Guided tends to produce more human-like and minimal edits that are not necessarily more faithful.

\begin{table}[h]
\centering
\small
\setlength{\tabcolsep}{4pt}
\begin{tabular}{llcc}
\hline
Dataset & Family & Faithfulness & ESMP \\
\hline
Movies & LLaMA & 0.042 $\rightarrow$ 0.893 & 0.334 $\rightarrow$ 0.500 \\
Movies & Qwen  & 0.124 $\rightarrow$ 0.873 & 0.298 $\rightarrow$ 0.585 \\
e-SNLI & LLaMA & 0.198 $\rightarrow$ 0.885 & 0.257 $\rightarrow$ 0.505 \\
e-SNLI & Qwen  & 0.514 $\rightarrow$ 0.936 & 0.341 $\rightarrow$ 0.498 \\
\hline
\end{tabular}
\caption{\textbf{Smallest-to-largest model comparison.} Values are averaged over prompting strategies and show how faithfulness and ESMP change from the smallest to the largest model within each dataset and model family.}
\label{tab:scale-comparison}
\end{table}

\end{document}